\documentclass[lettersize,journal]{IEEEtran}
\usepackage{amsmath,amsfonts}
\usepackage{array}
\usepackage[caption=false,font=normalsize,labelfont=sf,textfont=sf]{subfig}
\usepackage{textcomp}
\usepackage{stfloats}
\usepackage{url}
\usepackage{verbatim}
\usepackage{graphicx}

\usepackage{multirow}
\usepackage{marvosym}
\usepackage{algpseudocode}
\usepackage{xcolor}
\usepackage{courier}
\usepackage{indentfirst}
\usepackage{amsmath}
\usepackage{amssymb}
\usepackage{booktabs}
\usepackage{balance}
\usepackage[capitalize]{cleveref}
\usepackage[ruled]{algorithm2e}
\usepackage{amsmath}

\DeclareMathOperator*{\argmin}{arg\,min}

\def\etal{\emph{et al}.}

\def\BibTeX{{\rm B\kern-.05em{\sc i\kern-.025em b}\kern-.08em
    T\kern-.1667em\lower.7ex\hbox{E}\kern-.125emX}}

\begin{document}
\title{Distribution-Specific Learning for Joint Salient and Camouflaged Object Detection}

\author{
$\text{Chao Hao}$, 
$\text{Zitong Yu}$,~\IEEEmembership{Senior Member,~IEEE}, 
$\text{Xin Liu}$,~\IEEEmembership{Senior Member,~IEEE}, 
$\text{Yuhao Wang}$, 
$\text{Weicheng Xie}$,
 $\text{Jingang Shi}$, 
 Huanjing Yue,~\IEEEmembership{Senior Member,~IEEE}, 
$\text{Jingyu Yang}$,~\IEEEmembership{Senior Member,~IEEE}
\thanks{Received August, 2025. This work was supported by Guangdong Basic and Applied Basic Research Foundation (Grant No. 2023A1515140037). Chao Hao and Zitong Yu are co-first authors and they contribute equally. Corresponding authors: Xin Liu and Zitong Yu.}

\thanks{Chao Hao, Zitong Yu and Yuhao Wang are with the School of Computing and Information Technology, Great Bay University, Dongguan 523000, China (email: 3018234336@tju.edu.cn, yuzitong@gbu.edu.cn, wangyuhao@bupt.edu.cn).}

\thanks{Xin Liu is with Computer Vision and Pattern Recognition Laboratory, School of Engineering Science, Lappeenranta-Lahti University of Technology LUT, Lappeenranta 53850, Finland (e-mail: linuxsino@gmail.com).} 

\thanks{Weicheng Xie is with the School of Computer Science and Software Engineering, Shenzhen University, Shenzhen 518060, China (email: wcxie@szu.edu.cn).}

\thanks{Jingang Shi is with the School of Software Engineering, Xi'an Jiaotong University, Xi'an 710049, China (email: jingang.shi@hotmail.com).}

\thanks{Huanjing Yue and Jingyu Yang are with the School of Electrical and Information Engineering, Tianjin University, Tianjin 300072, China (email: huanjing.yue@tju.edu.cn, yjy@tju.edu.cn).}
}


\maketitle

\begin{abstract}
Salient object detection (SOD) and camouflaged object detection (COD) are two closely related but distinct computer vision tasks. Although both are class-agnostic segmentation tasks that map from RGB space to binary space, the former aims to identify the most salient objects in the image, while the latter focuses on detecting perfectly camouflaged objects that blend into the background in the image. These two tasks exhibit strong contradictory attributes. Previous works have mostly believed that joint learning of these two tasks would confuse the network, reducing its performance on both tasks. However, here we present an opposite perspective: with the correct approach to learning, the network can simultaneously possess the capability to find both salient and camouflaged objects, allowing both tasks to benefit from joint learning. We propose SCJoint, a joint learning scheme for SOD and COD tasks, assuming that the decoding processes of SOD and COD have different distribution characteristics. The key to our method is to learn the respective means and variances of the decoding processes for both tasks by inserting a minimal amount of task-specific learnable parameters within a fully shared network structure, thereby decoupling the contradictory attributes of the two tasks at a minimal cost. Furthermore, we propose a saliency-based sampling strategy (SBSS) to sample the training set of the SOD task to balance the training set sizes of the two tasks. In addition, SBSS improves the training set quality and shortens the training time. Based on the proposed SCJoint and SBSS, we train a powerful generalist network, named JoNet, which has the ability to simultaneously capture both “salient” and “camouflaged”. Extensive experiments demonstrate the competitive performance and effectiveness of our proposed method. The code is available at  \url{https://github.com/linuxsino/JoNet}.
\end{abstract}

\section{Introduction}
\label{sec:intro}

\IEEEPARstart{S}{alient} object detection (SOD) and camouflaged object detection (COD) are two similar yet significantly different tasks \cite{UJSC, UJSC-V2}. The similarity lies in that both are class-agnostic binary segmentation tasks, mapping RGB input images to single-channel binary images divided into target and non-target areas \cite{SENet}. The difference, however, lies in their focuses: the former concentrates on the most prominent object in the image \cite{SOD2}, while the latter focuses on inconspicuous objects camouflaged within their surroundings \cite{SINet}.

\begin{figure}[t]
      \centering
      \includegraphics[width=\linewidth]{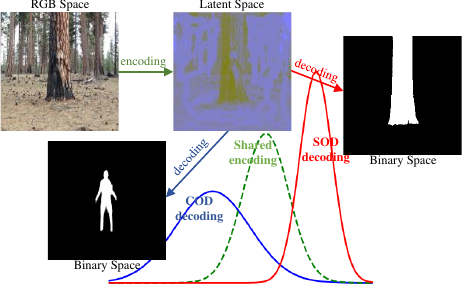}
      \caption{
       The distribution characteristics of the encoding process for the two tasks is unified, while that of decoding process is different. The concept of “distribution characteristics of the process” we propose here refers to the impact of the process on the data distribution.
      Given an image, our network has the capability to simultaneously capture both “salient” and “camouflaged”. }
      \label{fig:1}
\end{figure}

The emergence of many generalist models \cite{oneformer, EVP, generalist1, generalist2} recently has brought us numerous surprises. Generalist models have two main advantages: one is that they can complete multiple tasks simultaneously, avoiding the cumbersome need to repeatedly train several task-specific specialist models, thus enhancing training efficiency \cite{unifiedio}. Additionally, generalist models can benefit multiple tasks during joint training, further improving model performance and robustness \cite{mtl3}. Based on the introduction above, we know that the input and output formats of the SOD and COD tasks are entirely identical, and both share significant commonalities. Therefore, they are naturally suited to be processed with the same network architecture. This leads us to a natural idea: to jointly learn a generalist model for SOD and COD tasks.

In fact, there have been some works \cite{UJSC, UJSC-V2, SENet} involving the joint learning of SOD and COD before, but they did not achieve substantial success. Either the network structure is too complex, equivalent to multiple networks, or the performance is poor, not benefiting both tasks during joint training. Upon careful analysis, we find that these failures are mainly due to two primary challenges in the joint learning: 

\begin{figure}[t]
      \centering
      \includegraphics[width=0.96\linewidth]{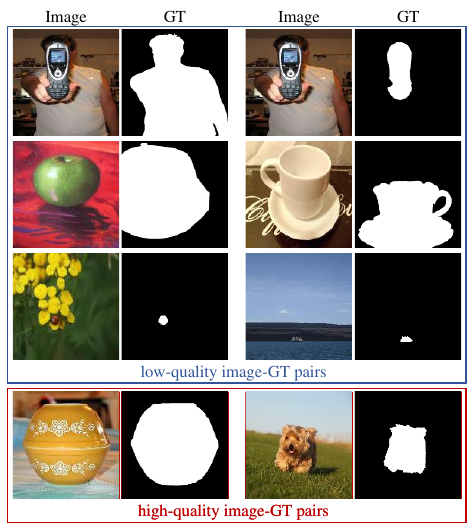}
       \caption{
      The first row: the same image appears twice in the training set with two different GTs. The second row: the GT does not match the original image. The third row: the target object given by the GT is not salient enough in the original image, inconsistent with the concept of “salient”.  And the fourth row: examples of high-quality image-GT pairs. The first three rows are selected from the lower ranks from SBSS, while the fourth row is selected from the higher ranks.}
      \label{fig:2}
\end{figure}

\begin{itemize}
\item The first and foremost challenge is the significant contradictory attributes between SOD and COD  tasks. Both are class-agnostic, treating the neural network as a black box with powerful fitting capabilities, the former requires the network to learn the abstract concept of “salient”, while the latter requires the network to learn the abstract concept of “camouflaged”. Therefore, \textbf{the joint learning of the two will bring confusion to the neural network}.

\item The second challenge is \textbf{the imbalance of training set sizes}. The SOD training set contains approximately $2.5$ times the number of images as the COD training set. It is known that when the sizes of the training sets for two tasks are inconsistent, the training of the network tends to be dominated by the one with the larger scale \cite{imbalance1, imbalance2, meng2025, arr}.
\end{itemize}

 For the first challenge, inspired by AdvProp \cite{advprop}, which hypothesizes that clean images and adversarial images belong to two different distributions, by using two batch norms to process the two types of data respectively, finally makes the network trained with adversarial samples and clear images to improve the recognition rate of clean images. We propose SCJoint, an extremely simple and effective joint learning scheme for SOD and COD tasks. As shown in \cref{fig:1}, the encoding process for the two tasks is unified, but the decoding processes are different. In the decoding process, we utilize two sets of different means and variances to change the distribution of the data, obtaining two types of outputs corresponding to “salient” and “camouflaged”, respectively. Therefore, we call that the two decoding processes as having different distribution characteristics, the concept of “distribution characteristics of the process” we propose here refers to the impact of the process on the data distribution. Specifically, we introduce two different task-specific distribution learning modules (DLM) in the decoder to learn their differences, while the rest of the network parameters are fully shared. We utilize a fully shared backbone network to learn the commonalities between the two tasks and use task-specific modules with very few parameters (negligible compared to shared parameters) to learn the unique characteristics of each task, enabling the network to simultaneously grasp the abstract concepts of “salient” and “camouflaged”.
 
 For the second challenge, we propose a saliency-based sampling strategy (SBSS) to clean the training set of SOD while reducing its size to be comparable to that of the COD training set. \cref{fig:2} is an example of low-quality image-GT pairs and high-quality image-GT pairs found using SBSS. A portion of the SOD training set is of low quality \cite{sod}, either because the objects in the images are not salient enough, or because there are issues with the corresponding ground truth (GT). We rank the image-GT pairs in the SOD training set based on saliency, with the higher-quality pairs ranking higher. We then select the top K pairs as our training dataset (in implementation, we choose K to be the number of images in the COD training set), eliminating those low-quality pairs that could potentially misguide the network. SBSS not only improves the quality of the SOD training set but also reduces its size to make it comparable to that of the COD training set, facilitating data balance in joint learning. Since the number of training images in the SOD task has been greatly reduced (the actual number has been reduced from 10553 to 4040), the training time of the network has been greatly reduced, almost half of that of independent training. The training set of the COD task is of high quality and the quantity is small and cannot be reduced, so we do not sample the COD training set. Note that the examples shown in \cref{fig:2} simply highlight that there are samples of lower quality in the SOD training set, not implying that the overall dataset quality is very poor, these lower-quality samples represent only a small fraction. There are quality variances among different image-GT pairs, and our emphasis is on SBSS's ability to detect such differences.

In summary, our main contributions include:
\begin{itemize}

\item We propose SCJoint, a very elegant joint learning scheme for SOD and COD. By introducing negligible extra parameters into the backbone network, enabling the network to simultaneously grasp the  “salient” and “camouflaged”.

\item We propose SBSS, which reduces the size and enhances the quality of the SOD training set by ranking based on saliency. This balances the scale of the SOD training set with the COD training set in joint learning, reducing training time and improving joint learning performance.

\item We achieve state-of-the-art (SOTA) performance on nine datasets of SOD and COD tasks, even outperforming those task-specific specialist models. Additionally, we conduct extensive experiments to demonstrate the effectiveness of the proposed approach.
\end{itemize}

 \begin{figure*}[t]
      \centering
      \includegraphics[width=0.96\linewidth]{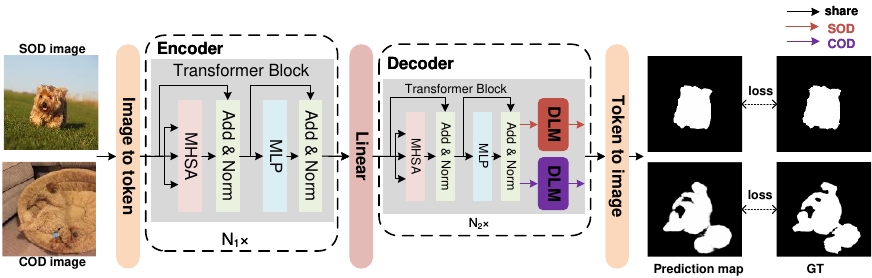}
      \caption{Illustration of the overall network architecture of our proposed SOD and COD joint training method. The entire network is shared, but we set up separate distribution learning modules (DLMs) for the two tasks in each Transformer block of the decoder (\textcolor[RGB]{192, 80, 70}{red} for SOD, \textcolor[RGB]{112, 48, 160}{purple} for COD). It is equivalent to our network having two modes. During training, SOD data uses SOD mode, and COD data uses COD mode. During inference, two predictions corresponding to “salient” and “camouflaged” can be generated for the same image. For the specific training process, see Algorithm 2.} 
      \label{fig:architecture}
\end{figure*}

\section{Related Work}
\label{sec:related}

\subsection{Salient and Camouflaged Object Detection}
\noindent \textbf{Salient object detection}. SOD has long been a hot task in computer vision field \cite{SOD1, SOD2, tifs3}. It is a classic binary segmentation task aimed at segmenting the most salient object in an image, regardless of the object's category \cite{MINet}. Since the advent of the deep learning era in computer vision, the field has seen rapid development \cite{SOD8, SOD10, SOD11, PDNet}. Neural networks, with their powerful fitting capabilities, can quickly capture the abstract concept of “salient” in a black-box manner, effectively accomplishing this task \cite{SOD4, SOD5, SOD6}. In the early stages, researchers primarily accomplished this task through the use of CNN-based Unet architectures \cite{MINet, SOD7}.
 Pang \etal \cite{MINet}  proposed the aggregate interaction modules to integrate the features from adjacent levels, in which less noise is introduced because of only using small up-/down-sampling rates. 
 Wei \etal \cite{SOD7} proposed the $F^3Net$ to solve the problem of differences in the receptive fields of different convolutional layers, which mainly consists of cross feature module and cascaded feedback decoder trained by minimizing a new pixel position aware loss.
Since the Transformer \cite{Transformer} was introduced into computer vision, the ViT-based \cite{ViT} encoder-decoder architecture has become mainstream. Liu \etal \cite{VST} proposed a new token upsampling method under the Transformer framework and fused multi-level patch tokens to better learn the details of the image. Liu \etal \cite{EVPv2} introduced a small number of learnable parameters into the Transformer architecture to learn task-specific knowledge, making full use of the powerful capabilities of the pre-trained model.

\vspace{1mm}
\noindent \textbf{Camouflaged object detection}. Different from SOD, the COD task emerged later. Fan \etal \cite{SINet} were the first to introduce the concept of camouflaged object detection and released the first large-scale COD dataset, establishing a benchmark. Upon its emergence, this task immediately garnered significant interest within the community \cite{ZoomNet, SegMAR, FEDER, tifs1, ease}. 
Due to the scarcity of camouflaged object images and the greater difficulty in annotation, the dataset size for this task is much smaller compared to SOD. Moreover, given the abstract concept of “camouflage”, although both tasks fall under class-agnostic segmentation, it is much more challenging for networks to detect camouflaged objects that are well-hidden within the background \cite{vscode, UJSC}. 
The existing mainstream methods are also based on the encoder-decoder architecture of Transformer. Huang \etal \cite{FSP} proposed a novel Transformer-based feature shrinkage pyramid network, which aims to hierarchically decode locality-enhanced neighboring transformer features through progressive shrinking to enhance local modeling and feature aggregation capabilities for camouflaged object detection. Lamdouar \etal \cite{cod1} addressed the question of what makes a camouflage successful, by proposing three scores for automatically assessing its effectiveness, through introducing a generative adversarial mechanism into the network, the network can excellently break camouflage. Zhou \etal \cite{tmm1} proposed a novel Decoupling and Integration Network (DINet) to detect camouflaged objects to solve many challenges such as low contrast between camouflaged objects and backgrounds and large changes in the appearance of camouflaged objects.

The mainstream methods for both SOD and COD tasks are based on the encoder-decoder architecture, which maps the RGB input to a latent space through the encoder and then maps the intermediate features from the latent space to a binary space output through the decoder \cite{sod12, sod13, cod10, cod11}. This general architecture has also laid the foundation for our work. We find that the encoding processes for the two tasks are similar, and previous work \cite{SENet} has also confirmed that sharing an encoder is feasible, the main difference between the two tasks lies in the differences in the decoding process.

\vspace{1mm}
\noindent \textbf{Joint learning}. Building a unified segmentation model is  currently a hot topic in the community, Some works \cite{oneformer, OMGSeg} have tried to use the same network to simultaneously handle semantic segmentation, instance segmentation, and panoptic segmentation tasks. For SOD and COD, two category-independent segmentation tasks, their joint learning is more difficult due to some potentially contradictory properties, but some works have explored it. Li \etal \cite{UJSC} started by exploiting the easy positive samples in the COD dataset to serve as hard positive samples in the SOD task to improve the robustness of the SOD model, using a shared decoder architecture to measure the similarity of two samples. But this is essentially data augmentation for the SOD task, and makes the network perform poorly on the COD task. Hao \etal \cite{SENet} used the shared encoder architecture to preliminarily explore the joint learning of SOD and COD in their work, but the problem of data imbalance between the two tasks also made their joint network perform poorly on the COD task. Luo \etal \cite{vscode} constructed different 2D prompts for SOD and COD tasks and different types of data to learn the differences between them, but they used too much data to achieve good performance, making the training cost too high and the training difficulty too great. In this work, we want to achieve better performance at a lower cost.

\subsection{Multitask Learning}
Multi-task learning generally refers to the use of a single network to accomplish multiple deep learning tasks \cite{mtl5, mtl6, mtl7, aud, auformer}, avoiding the cumbersome process of training several specialized models for specific tasks \cite{MTL1, MTL2}. It can save network parameters and improve training efficiency. Sometimes, through the joint learning of multiple tasks, it can even enhance the network's generalizability and performance on various tasks \cite{oneformer, EVP, unifiedio, xing2024emo, xing2025ttt}. Zhu \etal \cite{mtl3} greatly improved the accuracy of self-supervised depth estimation by simultaneously adding semantic segmentation of images to the self-supervised depth estimation task of images. Zeng \etal \cite{mtl4} added image reconstruction as an auxiliary task to the training process of image inpainting, which greatly improved the network's ability to perceive images and thus improved the performance of image inpainting.

There are still some differences between the joint learning of SOD and COD in our work and the common multi-task learning. Common multi-task learning usually involved a single sample corresponding to multiple labels for different tasks \cite{MTL1, mtl3, tifs5, tifs4, zhu, gui}, while we train using data from two tasks together, without any direct association between the two types of data. Additionally, previous multi-task learning often chose tasks that were very harmonious for joint learning \cite{mtl4, mtl7, tifs2}, with no apparent conflict between them. However, there are significant contradictory attributes between the SOD and COD tasks, which brings more challenges to our joint learning and increases the difficulty.

\section{Methodology}
\label{sec:method}
\subsection{Overview}
The overall network architecture of our joint learning for SOD and COD tasks is shown in \cref{fig:architecture}. We select SENet  \cite{SENet} as our backbone network (we only retain the SENet backbone part and do not use its LICM module) primarily due to its simple architecture, which is an asymmetric ViT-based \cite{ViT} encoder-decoder structure, containing only two ViTs without any additional complex designs. We provide more details about SENet in the supplementary material. Due to its simple architecture, it easily integrates with our joint training method (in fact, our approach can be combined with any network based on the Transformer \cite{Transformer} architecture). For ease of mention, we name our joint learning network for SOD and COD tasks as \textbf{JoNet}.

Specifically, the image is first divided into non-overlapping patches and tokenized to form tokens. The encoder then encodes these tokens, extracting useful features through self-attention mechanisms, the encoding process for both tasks is unified. However, during the decoding process, we apply different DLMs to learn two different decoding distribution characteristics for the two tasks (SOD DLM for SOD data, COD DLM for COD data). Finally, the tokens processed by decoder are used to generate a prediction map. This image-to-token modeling followed by token-to-image reconstruction pattern has been proven effective by MAE \cite{MAE}.

\subsection{Definition}
In this section, we introduce some notation definitions in this paper. We use $\mathcal{D}_{S}$ and  $\mathcal{D}_{C}$ to represent the training sets of the SOD and COD tasks respectively, \( \mathcal{D}_{S} = \{ (x_i, y_i) \}_{i=1}^{N_{S}} \) and \( \mathcal{D}_{C} = \{ (x_i, y_i) \}_{i=1}^{N_{C}} \), where $N_{S}$ and $N_{C}$ represent the number of images in $\mathcal{D}_{S}$ and  $\mathcal{D}_{C}$ respectively, $N_{S} > N_{C}$, $(x, y)$ represents the image-GT pair. Let $\mathcal{D}_{S}^{sub}$  represent the subset sampled from $\mathcal{D}_{S}$ using SBSS, \( \mathcal{D}_{S}^{sub} = \{ (x_i, y_i) \}_{i=1}^{N_{C}} \), we train the network using  $\mathcal{D}_{S}^{sub}$ and $\mathcal{D}_{C}$ together, both of size $N_{C}$. We use $\theta$ to represent the shared parameter part of the network, and $(\mu_s, \sigma_s)$, $(\mu_c, \sigma_c)$ to represent the two sets of task-specific parameters for COD and SOD respectively, where $\mu$ is the mean and $\sigma$ is the variance. Let $\mathcal{L}(.,.,.)$ to represent the loss function, $\mathcal{L}^S(\theta,(\mu_s, \sigma_s),(x^s,y^s))$ represents the  process of calculating loss for a SOD mini-batch $(x^s,y^s)$ and
$\mathcal{L}^C(\theta,(\mu_c, \sigma_c),(x^c,y^c))$ represents the  process of calculating loss for a COD mini-batch $(x^c,y^c)$.

\begin{figure}[t]
      \centering
      \includegraphics[width=\linewidth]{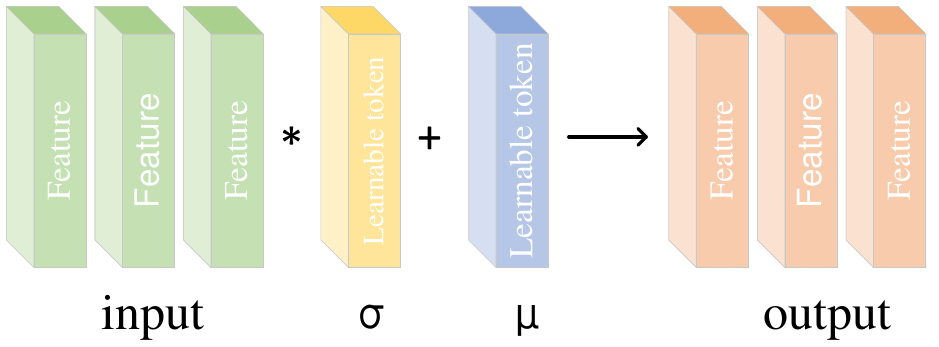}
      \caption{Illustration of the forward process of DLM. We set two learnable tokens to learn the mean and variance in the decoding process. The DLMs for SOD and COD are separate and have two different sets of parameters, $(\mu_s, \sigma_s)$ for SOD and $(\mu_c, \sigma_c)$ for COD.}
      \label{fig:DLM}
\end{figure}

\subsection{Saliency-based Sampling Strategy}
\label{method:sbss}
We formally propose SBSS in Algorithm 1 to filter out high-quality image-GT pairs in $\mathcal{D}_{S}$ and form a new subset $\mathcal{D}_{S}^{sub}$  for training. For each image $x_i$ in $\mathcal{D}_{S}$, we generate a prediction map $\hat{y}_i$ using SENet \cite{SENet} which has been trained on $\mathcal{D}_{S}$, and calculate the similarity between $\hat{y}_i$ and the corresponding GT $y_i$, the similarity between the two is quantified through the indicators introduced in \cref{eq:6}. We use this quantification result to represent the saliency of this image-GT pair, and select the highest-ranking $N_C$ pairs to form $\mathcal{D}_{S}^{sub}$ (choosing $N_C$ pairs is to make the training set sizes of the two tasks the same). 

We do this based on the fact that SENet trained on $\mathcal{D}_{S}$ has the ability to capture “salient”. If the prediction map of an image is more similar to its corresponding GT, it means that the quality of the image-GT pair is higher. We show in \cref{fig:2} the lower-ranked low-quality examples that we discarded and the high-quality examples that we retained in $\mathcal{D}_{S}^{sub}$.

\begin{algorithm}[t]
\DontPrintSemicolon
\KwData{$\mathcal{D}_{S}$}
\KwResult{$\mathcal{D}_{S}^{sub}$\\ }
\For{$(x_i, y_i)$ in $\mathcal{D}_{S}$}{
Use pretrained SOD network $f$ to generate a prediction map $\hat{y}_i$ for $x_i$, $\hat{y}_i = f(x_i);$\\
Use the similarity function $g$ to calculate the saliency score $S_i$ between $\hat{y}_i$ and $y_i$, $S_i = g(\hat{y}_i, y_i)$;\\
Use a dictionary $Dict$ to store corresponding image-GT pair $(x_i, y_i)$ and their saliency score $S_i$, $Dict[(x_i, y_i)] = S_i$.
}
Sort the dictionary $Dict$ according to value and take out the $N_C$ pairs with the highest $S_i$ to get $\mathcal{D}_{S}^{sub}$.\\
\KwRet{$\mathcal{D}_{S}^{sub}$}
\caption{Pseudo code of SBSS}
\label{algo:SBSS}
\end{algorithm}

Subsequent experiments also confirmed the effectiveness of SBSS. It has the advantages of improving the quality of the dataset, balancing the data size between tasks, and reducing the training time, thus further improving the performance of joint learning.

\subsection{SCJoint}
We formally propose SCJoint in Algorithm 2, a very elegant joint learning scheme for SOD and COD tasks. Specifically, we feed the SOD mini-batch and COD mini-batch to the same network but applied with different DLMs in the decoder for loss calculation, \textit{i.e.}, use the SOD DLM for SOD mini-batch and use COD DLM for COD mini-batch. The structure of DLM is shown in \cref{fig:DLM}, we set two different sets of learnable parameters to learn different distribution parameters in the two decoding processes, for SOD:
\begin{equation}
\label{eq:1}
X' = \dfrac{X-\mu_s}{\sqrt{\sigma_s^2}},
\end{equation}
and for COD:
\begin{equation}
\label{eq:2}
X' = \dfrac{X-\mu_c}{\sqrt{\sigma_c^2}},
\end{equation}
where $X'$ and $X$ represent output and input features respectively, $(\mu, \sigma)$ is used to learn the mean and variance of the distribution for the decoding process. The role of DLM here is somewhat akin to LayerNorm \cite{layernorm}, but LayerNorm is more to normalize the data to facilitate the stability of training. In addition, experiments have also revealed that employing DLM achieves better results.


\begin{algorithm}[t]
\DontPrintSemicolon
\KwData{$\mathcal{D}_{S}^{sub}$ and $\mathcal{D}_{C}$}
\KwResult{Shared network parameters $\theta$, task-specific parameters $(\mu_{s}, \sigma_{s})$ and  $(\mu_{c}, \sigma_{c})$;\\ }
\For{each training step}{
Sample a SOD image mini-batch $(x^s,y^s)$ from $\mathcal{D}_{S}^{sub}$;\\
Compute loss $\mathcal{L}^S(\theta, (\mu_{s}, \sigma_{s}),  (x^s, y^s))$ on SOD mini-batch $(x^s,y^s)$ using SOD DLM;\\
Sample a COD image mini-batch $(x^c,y^c)$ from $\mathcal{D}_{C}$;\\
Compute loss $\mathcal{L}^C(\theta, (\mu_{c}, \sigma_{c}),  (x^c, y^c))$ on COD mini-batch $(x^c,y^c)$ using COD DLM;\\
Minimize the total loss w.r.t. $\theta$,  $(\mu_{s}, \sigma_{s})$ and $(\mu_{c}, \sigma_{c})$: $\argmin \limits_{\theta, (\mu_{s}, \sigma_{s}), (\mu_{c}, \sigma_{c})} [\mathcal{L}^S(\theta,(\mu_{s}, \sigma_{s}), (x^s, y^s)) + \mathcal{L}^C(\theta,(\mu_{c}, \sigma_{c}), (x^c, y^c))$].
}
\KwRet{$\theta$, $(\mu_{s}, \sigma_{s})$ and $(\mu_{c}, \sigma_{c})$}
\caption{Pseudo code of SCJoint}
\label{algo:SCJoint}
\end{algorithm}

For SOD mini-batch $(x^{s}, y^{s})$, the objective function is 
\begin{equation}
\label{eq:3}
\argmin \limits_{\theta, (\mu_{s}, \sigma_{s})} \mathcal{L}^S(\theta,(\mu_{s}, \sigma_{s}), (x^s, y^s)),
\end{equation}
the SOD task will affect the updates of $\theta$ and $(\mu_{s}, \sigma_{s})$.

For COD mini-batch $(x^{c}, y^{c})$, the objective function is
\begin{equation}
\label{eq:4}
\argmin \limits_{\theta, (\mu_{c}, \sigma_{c})} \mathcal{L}^C(\theta,(\mu_{c}, \sigma_{c}), (x^c, y^c)),
\end{equation} 
the COD task will affect the updates of $\theta$ and $(\mu_{c}, \sigma_{c})$.

Therefore, for the joint learning, the total objective function is 
\begin{equation}
\label{eq:5}
\argmin \limits_{\theta, (\mu_{s}, \sigma_{s}), (\mu_{c}, \sigma_{c})} \mathcal{L}^S(\theta,(\mu_{s}, \sigma_{s}), (x^s, y^s)) + \mathcal{L}^C(\theta,(\mu_{c}, \sigma_{c}), (x^c, y^c)).
\end{equation} 

According to \cref{eq:3}, \cref{eq:4} and \cref{eq:5}, it can be clearly seen that except DLM, all the remaining parameters $\theta$ are jointly optimized for both SOD samples and COD samples, $(\mu_{s}, \sigma_{s})$ is optimized only by SOD samples and $(\mu_{c}, \sigma_{c})$ is optimized only by COD samples. This is consistent with our idea of using a shared network structure to learn commonalities between two tasks and task-specific parameters to learn their respective characteristics.

It is necessary to add that we use Pixel Position Aware loss \cite{SOD7} to compute the segmentation loss for both SOD and COD tasks, given the prediction map $x$ and the ground truth $y$, the loss is defined as:
\begin{equation}
\label{eq:bceiou}
\mathcal{L}(x, y) = L_{BCE}^\omega(x, y) + L_{IOU}^\omega(x, y),
\end{equation} 
where $L_{BCE}^\omega$ and $L_{IOU}^\omega$ are weighted BCE and IOU loss. The above $\mathcal{L}^S$ and $\mathcal{L}^C$ are both calculated using \cref{eq:bceiou}.

Note the introduction of DLM in SCJonit only increases a negligible amount of extra parameters for network training based on the baseline model SENet, \textit{i.e.}, the four learnable vectors shown in \cref{fig:DLM}, with a total of about 2K parameters in the setting of this paper. During training, SOD data can only use SOD mode (using SOD DLM), and COD data can only use COD mode (using COD DLM), but in the inference phase, one image can get output in both modes. Furthermore, compared to the prompt learning method \cite{vscode}, our SCJoint also does not need to use additional losses to distinguish the two modes and is more logically and mathematically sound.

We used an extremely simple method to effectively complete the joint learning of SOD and COD. The experiments also show the effectiveness of our disentanglement strategy that divides joint learning into learning commonalities and characteristics.

\begin{table*}[t]
\centering
\caption{We select two SOD datasets DUTS \cite{DUTS} and DUT-OMRON \cite{DUT-OMRON} and two COD datasets CAMO \cite{CAMO} and COD10K \cite{SINet} to conduct ablation experiments to verify the effectiveness of our proposed SCJoint and SBSS. “IT” indicates independent training, \textit{i.e.}, use the SOD training set (the original dataset without the SBSS sampling strategy) or COD training set to train the network separately. “JT” indicates joint training, \textit{i.e.}, use the SOD and COD training set to train the network together. “$\uparrow / \downarrow$”: the higher/lower the better.}

\resizebox{\linewidth}{!}{\begin{tabular}{cccccccccccccccccc}
\hline
\multicolumn{1}{c|}{\multirow{2}{*}{Settings}} 
& \multicolumn{4}{c|}{DUTS}                    
& \multicolumn{4}{c|}{PASCAL-S}                    
& \multicolumn{4}{c|}{CAMO}                  
& \multicolumn{4}{c}{COD10K}      \\ \cline{2-17} 
\multicolumn{1}{c|}{}              

&$S_{\alpha}\uparrow$ &$E_{\phi}\uparrow$  &$F_{\beta}^{m}\uparrow$  &\multicolumn{1}{c|}{$M \downarrow$}       

&$S_{\alpha}\uparrow$ &$E_{\phi}\uparrow$  &$F_{\beta}^{m}\uparrow$  &\multicolumn{1}{c|}{$M \downarrow$}     

&$S_{\alpha}\uparrow$ &$E_{\phi}\uparrow$  &$F_{\beta}^{\omega}\uparrow$  &\multicolumn{1}{c|}{$M \downarrow$}   

&$S_{\alpha}\uparrow$ &$E_{\phi}\uparrow$  &$F_{\beta}^{\omega}\uparrow$    & $M \downarrow$     \\ \hline

\multicolumn{1}{l|}{IT}               
&.921  &.953  &.921  & \multicolumn{1}{c|}{.024}  
&.883  &.920  &.881 &\multicolumn{1}{c|}{.049} 
&.875  &.916  &.832 & \multicolumn{1}{c|}{.043} 
&.865  &.920  &.773  &.026  \\ \hline
 
\multicolumn{1}{l|}{JT}              
&.928  &.956  &.927 & \multicolumn{1}{c|}{.020} 
&.884  &.921  &.882 & \multicolumn{1}{c|}{.047} 
&.853  &.907  &.808 & \multicolumn{1}{c|}{.052} 
&.841  &.901  &.739 &.032  \\

\multicolumn{1}{l|}{JT + SBSS}               
&.928  &.958  &.925 & \multicolumn{1}{c|}{.021} 
&.887  &.924  &.883  & \multicolumn{1}{c|}{.046} 
&.862  &.911  &.813 & \multicolumn{1}{c|}{.048} 
&.857  &.918  &.765  &.028  \\

\multicolumn{1}{l|}{JT + SCJoint}               
&.927  &.954  &.927 & \multicolumn{1}{c|}{.023}  
&.886  &.921  &.891 & \multicolumn{1}{c|}{.050} 
&.873  &.919  &.833 & \multicolumn{1}{c|}{.045}  
&.864  &.922  &.777  &.025  \\  \hline

\multicolumn{1}{l|}{\textbf{JT + SBSS + SCJoint}}              
&\textbf{.931}  &\textbf{.958}  &\textbf{.929} & \multicolumn{1}{c|}{\textbf{.020}}  
&\textbf{.887}  &\textbf{.924}  &\textbf{.885} & \multicolumn{1}{c|}{\textbf{.046}} 
&\textbf{.889}  &\textbf{.931}  &\textbf{.847} & \multicolumn{1}{c|}{\textbf{.038}} 
&\textbf{.871}  &\textbf{.925}  &\textbf{.783}  &\textbf{.023}  \\
\hline

\end{tabular}}
\label{tab:efficacy}
\end{table*}

\begin{table*}[t]
\centering
\caption{Ablation study on locations of DLM. “encoder” indicates inserting DLM only in encoder. “decoder” indicates inserting DLM only in decoder, which is also our default setting in this work. “encoder and decoder” indicates inserting DLM in both encoder and decoder. “last x layers in decoder” indicates inserting DLM in the last x Transformer blocks of the decoder (8 in total).}%

\resizebox{\linewidth}{!}{\begin{tabular}{cccccccccccccccccc}
\hline
\multicolumn{1}{c|}{\multirow{2}{*}{Settings}} 
& \multicolumn{4}{c|}{DUTS}                    
& \multicolumn{4}{c|}{PASCAL-S}                    
& \multicolumn{4}{c|}{CAMO}                  
& \multicolumn{4}{c}{COD10K}      \\ \cline{2-17} 
\multicolumn{1}{c|}{}              

&$S_{\alpha}\uparrow$ &$E_{\phi}\uparrow$  &$F_{\beta}^{m}\uparrow$  &\multicolumn{1}{c|}{$M \downarrow$}       

&$S_{\alpha}\uparrow$ &$E_{\phi}\uparrow$  &$F_{\beta}^{m}\uparrow$  &\multicolumn{1}{c|}{$M \downarrow$}     

&$S_{\alpha}\uparrow$ &$E_{\phi}\uparrow$  &$F_{\beta}^{\omega}\uparrow$  &\multicolumn{1}{c|}{$M \downarrow$}   

&$S_{\alpha}\uparrow$ &$E_{\phi}\uparrow$  &$F_{\beta}^{\omega}\uparrow$    & $M \downarrow$     \\ \hline

\multicolumn{1}{l|}{encoder}               
&.877  &.909  &.855 & \multicolumn{1}{c|}{.041}  
&.827  &.860  &.812 & \multicolumn{1}{c|}{.083} 
&.880  &.920  &.835 & \multicolumn{1}{c|}{.042} 
&.866  &.919  &.775  &.026  \\
 
\multicolumn{1}{l|}{\textbf{decoder}}              
&\textbf{.931}  &\textbf{.958}  &\textbf{.929} & \multicolumn{1}{c|}{\textbf{.020}}  
&\textbf{.887}  &\textbf{.924}  &\textbf{.885} & \multicolumn{1}{c|}{\textbf{.046}} 
&\textbf{.889}  &\textbf{.931}  &\textbf{.847} & \multicolumn{1}{c|}{\textbf{.038}} 
&\textbf{.871}  &\textbf{.925}  &\textbf{.783}  &\textbf{.023}  \\

\multicolumn{1}{l|}{encoder and decoder}               
&.854  &.887  &.820 & \multicolumn{1}{c|}{.052} 
&.812  &.847  &.794 & \multicolumn{1}{c|}{.092} 
&.879  &.920  &.837 & \multicolumn{1}{c|}{.042} 
&.867  &.922  &.777  &.025  \\
\hline

\multicolumn{1}{l|}{last 1 layers in decoder}               
&.927  &.954  &.924   & \multicolumn{1}{c|}{.023}  
&.879  &.915  &.873   & \multicolumn{1}{c|}{.049} 
&.875  &.923  &.829 & \multicolumn{1}{c|}{.044}  
&.866  &.917  &.777  &.026  \\

\multicolumn{1}{l|}{last 2 layers in decoder}               
&.927  &.956  &.925 & \multicolumn{1}{c|}{.023}  
&.881  &.916  &.873 & \multicolumn{1}{c|}{.050} 
&.879 &.925  &.831 & \multicolumn{1}{c|}{.042}  
&.868  &.919  &.778  &.025  \\

\multicolumn{1}{l|}{last 4 layers in decoder}              
&.929  &.956  &.927 & \multicolumn{1}{c|}{.022}  
&.882  &.916  &.879 & \multicolumn{1}{c|}{.048} 
&.880  &.927  &.838 & \multicolumn{1}{c|}{.040} 
&.870  &.922  &.780  &.024  \\

\multicolumn{1}{l|}{last 6 layers in decoder}              
&.929  &.958  &.928 & \multicolumn{1}{c|}{.021}  
&.883  &.924  &.881 & \multicolumn{1}{c|}{.046} 
&.885  &.930  &.845 & \multicolumn{1}{c|}{.040} 
&.872  &.924  &.781  & .024 \\
\hline
 
 
\end{tabular}}
\label{tab:DLM_location}
\end{table*}

\section{Experiments}
\label{sec:experiment}
 \subsection{Experimental Setup}
\label{exp:setup}
 \textbf{Datasets}. For SOD task, we use five datasets used in most work: DUTS \cite{DUTS}, DUT-OMRON \cite{DUT-OMRON}, HKU-IS \cite{HKU-IS}, ECSSD \cite{ECSSD} and PASCAL-S \cite{PASCAL-S}, where DUTS provides $10553$ image-GT pairs for training, \textit{i.e.}, \( \mathcal{D}_{S} = \{ (x_i, y_i) \}_{i=1}^{10553} \). 
 
 For COD task, we use four datasets: CAMO \cite{CAMO}, 
 CHAMELEON \cite{CHAMELEON}, COD10K \cite{SINet} and NC4K \cite{NC4K}, where CAMO and COD10K provide $4040$ image-GT pairs for training, \textit{i.e.}, \( \mathcal{D}_{C} = \{ (x_i, y_i) \}_{i=1}^{4040} \). We use SBSS to sample $4040$ iamge-GT pairs from $\mathcal{D}_{S}$ to get $\mathcal{D}_{S}^{sub}$ and use $\mathcal{D}_{S}^{sub}$ and $\mathcal{D}_{C}$ as our training set in this work.

\vspace{1mm}
 \textbf{Metrics}. Structure measure  ($S_{\alpha}$), mean E-measure ($E_{\phi}$), F-measure ($F_{\beta}$), and mean absolute error ($M$) are reported to evaluate the performance. Note that weighted F-measure ($F_{\beta}^{\omega}$) is used for COD and maximum F-measure ($F_{\beta}^{m}$) is used for SOD. In addition, we use the following metric \cite{SENet}  to quantitatively calculate the similarity covered in \cref{method:sbss} (the higher, the better):
\begin{equation}
\label{eq:6}
\textit{S} = S_{\alpha} + E_{\phi} + F_{\beta} + (1-M) .
\end{equation}

 \textbf{Implementation details}. Our experiments are implemented in PyTorch on two 3090 GPUs and are  optimized by the Adam.  We resize all the input images to $384 \times 384$ and augment them by randomly horizontal flipping. We set the mini-batch size to 16, all parameters are updated during the training process of $25$ epochs, the learning rate is initialized to $10^{-4}$ and adjusted by poly strategy with the power of $0.9$. We do not initialize JoNet using SENet's \cite{SENet} weights. Instead, following its initial setup, we use pre-trained parameters from MAE \cite{MAE} to initialize both the encoder and decoder.

\begin{table*}[t]
\centering
\caption{Ablation study on the sampling part of SBSS. “Top-K” indicates sampling the image-GT pairs with the high ranking in saliency calculation, which is also our default setting in this work. “Bottom-K” indicates sampling the image-GT pairs with the low ranking in saliency calculation. “Random-K” indicates random sampling.}
\resizebox{\linewidth}{!}{\begin{tabular}{cccccccccccccccccc}
\hline
\multicolumn{1}{c|}{\multirow{2}{*}{Settings}} 
& \multicolumn{4}{c|}{DUTS}                    
& \multicolumn{4}{c|}{PASCAL-S}                    
& \multicolumn{4}{c|}{CAMO}                  
& \multicolumn{4}{c}{COD10K}      \\ \cline{2-17} 
\multicolumn{1}{c|}{}              

&$S_{\alpha}\uparrow$ &$E_{\phi}\uparrow$  &$F_{\beta}^{m}\uparrow$  &\multicolumn{1}{c|}{$M \downarrow$}       

&$S_{\alpha}\uparrow$ &$E_{\phi}\uparrow$  &$F_{\beta}^{m}\uparrow$  &\multicolumn{1}{c|}{$M \downarrow$}     

&$S_{\alpha}\uparrow$ &$E_{\phi}\uparrow$  &$F_{\beta}^{\omega}\uparrow$  &\multicolumn{1}{c|}{$M \downarrow$}   

&$S_{\alpha}\uparrow$ &$E_{\phi}\uparrow$  &$F_{\beta}^{\omega}\uparrow$    & $M \downarrow$     \\ \hline
\multicolumn{1}{l|}{\textbf{Top-K}}              
&\textbf{.931}  &\textbf{.958}  &\textbf{.929} & \multicolumn{1}{c|}{\textbf{.020}}  
&\textbf{.887}  &\textbf{.924}  &\textbf{.885} & \multicolumn{1}{c|}{\textbf{.046}} 
&\textbf{.889}  &\textbf{.931}  &\textbf{.847} & \multicolumn{1}{c|}{\textbf{.038}} 
&\textbf{.871}  &\textbf{.925}  &\textbf{.783}  &\textbf{.023}  \\
 
\multicolumn{1}{l|}{Bottom-K}              
&.919  &.948  &.920 & \multicolumn{1}{c|}{.026} 
&.875  &.911  &.877 & \multicolumn{1}{c|}{.051} 
&.881  &.925  &.837 & \multicolumn{1}{c|}{.042} 
&.867  &.923  &.778  &.026  \\

\multicolumn{1}{l|}{Random-K}               
&.926  & 954 &.921 & \multicolumn{1}{c|}{.023} 
&.880  &.917  &.881 & \multicolumn{1}{c|}{.048} 
&.886  &.930  &.841 & \multicolumn{1}{c|}{.039} 
&.870  &.922  &.781  &.024  \\

\hline
 
\end{tabular}}
\label{tab:sbss}
\end{table*}
\begin{table*}[t]
\centering
\caption{We select two SOD datasets DUTS \cite{DUTS} and ECSSD \cite{ECSSD} and two COD datasets CAMO \cite{CAMO} and NC4K \cite{NC4K} to observe the impact of training set size on performance for two tasks during joint training. We use all $4040$ COD training images and different numbers of SOD training images mixed together to train the network.}
\resizebox{\linewidth}{!}{\begin{tabular}{cccccccccccccccccc}
\hline
\multicolumn{1}{c|}{\multirow{2}{*}{Settings}} 
& \multicolumn{4}{c|}{DUTS}                    
& \multicolumn{4}{c|}{ECSSD}                    
& \multicolumn{4}{c|}{CAMO}                  
& \multicolumn{4}{c}{NC4K}      \\ \cline{2-17} 
\multicolumn{1}{c|}{}              

&$S_{\alpha}\uparrow$ &$E_{\phi}\uparrow$  &$F_{\beta}^{m}\uparrow$  &\multicolumn{1}{c|}{$M \downarrow$}       

&$S_{\alpha}\uparrow$ &$E_{\phi}\uparrow$  &$F_{\beta}^{m}\uparrow$  &\multicolumn{1}{c|}{$M \downarrow$}     

&$S_{\alpha}\uparrow$ &$E_{\phi}\uparrow$  &$F_{\beta}^{\omega}\uparrow$  &\multicolumn{1}{c|}{$M \downarrow$}   

&$S_{\alpha}\uparrow$ &$E_{\phi}\uparrow$  &$F_{\beta}^{\omega}\uparrow$    & $M \downarrow$     \\ \hline

\multicolumn{1}{l|}{1000}               
&.880 &.901  &.861  & \multicolumn{1}{c|}{.046}  
&.908  &.921  &.915  & \multicolumn{1}{c|}{.045}  
&.873  &.910  &.825  & \multicolumn{1}{c|}{.043}  
&.888  &.931  &.840  &.033  \\ 

\multicolumn{1}{l|}{2000}              
&.891  &.922  &.888  & \multicolumn{1}{c|}{.035}  
&.925  &.945  &.932  & \multicolumn{1}{c|}{.031}  
&.867  &.908  &.821  & \multicolumn{1}{c|}{.045}  
&.885  &.929  &.838  &.034  \\ 

\multicolumn{1}{l|}{4040}               
&.902  &.936  &.901  & \multicolumn{1}{c|}{.031}  
&.935  &.955  &.945  & \multicolumn{1}{c|}{.027}  
&.860  &.905  &.813  & \multicolumn{1}{c|}{.050}  
&.882  &.925  &.838  &.033  \\ 
 
\multicolumn{1}{l|}{8000}               
&.925  &.945  &.920  & \multicolumn{1}{c|}{.025}  
&.940  & 959 &.948 & \multicolumn{1}{c|}{.025}  
&.855  &.902  &.810  & \multicolumn{1}{c|}{.052}  
&.881  &.920  &.828  &.035  \\

\multicolumn{1}{l|}{10553}               
&.930  &.955  &.929 & \multicolumn{1}{c|}{.020}  
&.944  &.964  &.957  & \multicolumn{1}{c|}{.023}  
&.853 &.901  &.808  & \multicolumn{1}{c|}{.052}  
&.874  &.919  &.825  &.039  \\ \hline

\end{tabular}}
\label{tab:balance}
\end{table*}

\subsection{Ablation Study}
 \textbf{Validity of each part}. To demonstrate the effectiveness of each component in this work, we report the quantitative results in \cref{tab:efficacy}, we use the backbone of SENet \cite{SENet} as our baseline model and select 4 testing sets to report performance. We first report the results of independent training (IT) of the two tasks, training SENet on the training sets of the two tasks respectively and testing on the corresponding test sets. Next we report the joint training (JT) results without using any strategy, just mixing the training sets of the two tasks together. It can be seen that this simple joint training has a great negative impact on the performance of the COD task and has a slight positive impact on SOD. We think this is reasonable because the size of the SOD training set is much larger than that of COD. At this time, the network is led by SOD. This also aligns with the viewpoint mentioned in \cite{UJSC-V2} that COD samples are the more challenging SOD samples, and their joint training slightly benefits SOD.

\begin{table*}[t]
\centering
\caption{Quantitative comparison of our SCJoint with other joint learning methods. In order to make the comparison as fair as possible, we uniformly use SENet as the backbone network and use SBSS during training. “IT” indicates independent training. “Encoder sharing” refers to a setup where two tasks share a single encoder but each has its own dedicated decoder. “Decoder sharing” refers to the opposite scenario, where two tasks share a single decoder but each is equipped with a separate encoder. “Prompt learning” refers to adding learnable task-specific tokens for the two tasks in joint training. “SCJoint (LN)”  refers to the architecture in which LayerNorm is used to replace the DLM module.}
\resizebox{\linewidth}{!}{\begin{tabular}{cccccccccccccccccc}
\hline
\multicolumn{1}{c|}{\multirow{2}{*}{Settings}} 
& \multicolumn{4}{c|}{DUTS}                    
& \multicolumn{4}{c|}{PASCAL-S}                    
& \multicolumn{4}{c|}{CAMO}                  
& \multicolumn{4}{c}{COD10K}      \\ \cline{2-17} 
\multicolumn{1}{c|}{}              

&$S_{\alpha}\uparrow$ &$E_{\phi}\uparrow$  &$F_{\beta}^{m}\uparrow$  &\multicolumn{1}{c|}{$M \downarrow$}       

&$S_{\alpha}\uparrow$ &$E_{\phi}\uparrow$  &$F_{\beta}^{m}\uparrow$  &\multicolumn{1}{c|}{$M \downarrow$}     

&$S_{\alpha}\uparrow$ &$E_{\phi}\uparrow$  &$F_{\beta}^{\omega}\uparrow$  &\multicolumn{1}{c|}{$M \downarrow$}   

&$S_{\alpha}\uparrow$ &$E_{\phi}\uparrow$  &$F_{\beta}^{\omega}\uparrow$    & $M \downarrow$     \\ \hline

\multicolumn{1}{l|}{IT}               
&.921  &.953  &.921  & \multicolumn{1}{c|}{.024}  
&.883  &.920  &.881 &\multicolumn{1}{c|}{.049} 
&.875  &.916  &.832 & \multicolumn{1}{c|}{.043} 
&.865  &.920  &.773  &.026  \\ \hline

\multicolumn{1}{l|}{Encoder sharing \cite{SENet}}              
&.929  &.957  &.929  & \multicolumn{1}{c|}{.021}  
&.887  &.921  &.882  & \multicolumn{1}{c|}{.047}  
&.881  &.924  &.839  & \multicolumn{1}{c|}{.041}  
&.865  &.914  &.771  &.025  \\ 

\multicolumn{1}{l|}{Decoder sharing \cite{UJSC}}               
&.908  &.925  &.899  & \multicolumn{1}{c|}{.032}  
&.852  &.901  &.849  & \multicolumn{1}{c|}{.062}  
&.849  &.901  &.796  & \multicolumn{1}{c|}{.055}  
&.828  &.883  &.722  &.035  \\ 
 
\multicolumn{1}{l|}{Prompt learning \cite{vscode}}               
&.926  &.956  &.927  & \multicolumn{1}{c|}{.023}  
&.882  &.919  &.886  & \multicolumn{1}{c|}{.050}  
&.878  &.923  &.839  & \multicolumn{1}{c|}{.044}  
&.858  &.918  &.768  &.026  \\ \hline

\multicolumn{1}{l|}{SCJoint (LN)}               
&.923  &.954 &.928 & \multicolumn{1}{c|}{.024}  
&.881  &.920 &.880 & \multicolumn{1}{c|}{.047}
&.880  &.920  &.831 & \multicolumn{1}{c|}{\textbf{.038}}
&.865 &.919 &.775  &.026  \\

\multicolumn{1}{l|}{\textbf{SCJoint (Ours)}}               
&\textbf{.931}  &\textbf{.958}  &\textbf{.929} & \multicolumn{1}{c|}{\textbf{.020}}  
&\textbf{.887}  &\textbf{.924}  &\textbf{.885} & \multicolumn{1}{c|}{\textbf{.046}} 
&\textbf{.889}  &\textbf{.931}  &\textbf{.847} & \multicolumn{1}{c|}{\textbf{.038}} 
&\textbf{.871}  &\textbf{.925}  &\textbf{.783}  &\textbf{.023}  \\
\hline

\end{tabular}}
\label{tab:joint}
\end{table*}
\begin{table*}[t]
\centering
\caption{Quantitative comparison of our method with  other SOTA methods on five SOD benchmark datasets. “-”: Not available. The top two results are highlighted in \textcolor{red}{red} and \textcolor{blue}{blue}. }
\resizebox{\linewidth}{!}{\begin{tabular}{c|c|cccc|cccc|cccc|cccc|cccc}
\hline
\multirow{2}{*}{Method} & \multirow{2}{*}{Venue} 
& \multicolumn{4}{c|}{DUTS}                          
& \multicolumn{4}{c|}{DUT-OMRON}                     
& \multicolumn{4}{c|}{HKU-IS}                        
& \multicolumn{4}{c|}{ECSSD}                          
& \multicolumn{4}{c}{PASCAL-S}  \\ \cline{3-22} 
                        &        
&$S_{\alpha}\uparrow$ &$E_{\phi}\uparrow$  &$F_{\beta}^{m}\uparrow$  &$M \downarrow$ 
&$S_{\alpha}\uparrow$ &$E_{\phi}\uparrow$  &$F_{\beta}^{m}\uparrow$  &$M \downarrow$ 
&$S_{\alpha}\uparrow$ &$E_{\phi}\uparrow$  &$F_{\beta}^{m}\uparrow$  &$M \downarrow$ 
&$S_{\alpha}\uparrow$ &$E_{\phi}\uparrow$  &$F_{\beta}^{m}\uparrow$  &$M \downarrow$ 
&$S_{\alpha}\uparrow$ &$E_{\phi}\uparrow$  &$F_{\beta}^{m}\uparrow$  &$M \downarrow$ \\ 
\hline

F3Net \cite{SOD7}  &AAAI 20
&.888 &.902 &.840 &.035
&.838 &.870 &.766 &.053
&.917 &.953 &.910 &.028
&.924 &.927 &.925 &.033
&.855 &.859 &.840 &.062  \\

MINet \cite{MINet}  &CVPR 20
&.884 &.917 &.884 &.037
&.833 &.873 &.810 &.055
&.920 &.961 &.935 &.028
&.925 &.953 &.947 &.033
&.857 &.899 &.882 &.064  \\

LDF \cite{LDF}                      &   CVPR 20             
&  .892     &  .925     & .861      &   .034 
&   .839     & .865      & .770      &  .052  
&   .920    &  .953     &   .913    &  .028  
& .919      & .943      & 923      &  .036     
&  .860     & .901      &  .856     & .063      \\

SINet \cite{SINet}       &CVPR 20
&.872    &.904    &.846    &.042
&.825    &.851    &.757    &.058
&.911    &.942    &.916    &.033
&.916    &.939    &.929    &.041
& .855   &.888    &.842    &.069     \\                         

BGNet \cite{BGNet} &IJCAI 21
&.891 &.922 &.873 &.033
&.834 &.858 &.769 &.050
&.917 &.951 &.926 &.028
&.930 &.954 &.942 &.029
&.862 &.900 &.849 &.058 \\

VST \cite{VST}                      & ICCV 21         
& .896      & .892      & .890      &  .037    
&  .850      & .861      & .825      & .058     
&  .928     &.953       & .942      &  .029    
&  .932     & .918      &  .951     & .033     
& .865      & .837      &  .875     & .061      \\

UJSC \cite{UJSC}    & CVPR 21           
& .899      &  .937     & .866      & .032      
& .850       & .884      &  .782     &  .051    
& .931     &  \textcolor{red}{.967}     & .924      & .026     
& .933     &  .960     & .935      & .030     
& .864    & .902      &  .841     &  .062     \\

SelfReformer \cite{SelfReformer}              &   TMM 22                   
& .911      & .920      &  .916     &  .026    
&  .856      &  .886     & .836      &  \textcolor{blue}{.041}    
&  .930     & .959      & .947      & .024   
&   .935    &  .928     &  .957     &  .027    
&  .874     & .872      &  \textcolor{red}{.894}     &  .050     \\

UPL \cite{UPL}                       &  AAAI 22                     
& .846      & -      &.783       & .050     
&  .808      &  -     & .711      & .059     
& .897      &-       & .879      & .035     
& .899      & -      & .885      & .043     
& .822      &  -     &  .773     &   .080    \\

PSOD \cite{PSOD}                       &  AAAI 22                
&  .853     & -      & .858      &  .045  
&  .824      & -      & .809      &   .064 
&  .902     &  -     &   .923    &   .032  
&  .914     & -      &   .936    &   .035 
& .853      & -      &   .866    &  .064     \\

ZoomNet \cite{ZoomNet}    &CVPR 22
&.900 &.936 &.866 &.033
&.841 &.872 &.771 &.053
&.931 &\textcolor{red}{.967} &.923 &.023
&.935 &.963 &.933 &.027
&.869 &\textcolor{blue}{.917} &.860 &.057   \\

BBRF \cite{BBRF}                 &     TIP 23                 
& .908      &  .927     & .916      &   \textcolor{blue}{.025 }   
& .855       & .887      & .843      &  .042    
& .935      &  .965     &  \textcolor{red}{.958}     & \textcolor{blue}{ .020 }   
&  .939     & .934      &  .963     & \textcolor{blue}{ .022}     
&  .871     &  .867     &  \textcolor{blue}{.891}     & \textcolor{blue}{ .049}    \\

TGSD \cite{TGSD}                      &CVPR 23               
&.842       & .902      & .824      &   .047
& .812       &.863       & .763      &   .061 
&  .890     & .942      &  .906     &    .036 
&  .894     &  .937     &  .918     &  .044  
&   .830    & .887      & .828      &   .072    \\

EVP \cite{EVP}     &   CVPR 23                    
& .913 & .947 &\textcolor{blue}{.923} & .026 
& .862  & .894 & .858 & .046 
& .931 & .961 & .952 & .024 
& .935 & .957 & .960 & .027 
& \textcolor{blue}{.878} & \textcolor{blue}{.917} & .872 & .054 \\

UJSC-V2 \cite{UJSC-V2}             & arXiv 23          
&  .900     & .937      &  .875     & .030
&  .841     & .876      & .777       &.050 
& .921     &.958       & .920     & .026  
&   .929    & .955      &.935      &  .029  
&   .866    &.910     & .867      &  .058     \\

EVP-V2 \cite{EVPv2}             & arXiv 23          
&  .915    & .944      &  .921     & .026

& \textcolor{red}{.874}     & \textcolor{red}{.902}      & \textcolor{red}{.865}  &\textcolor{red}{.040} 

&  \textcolor{blue}{.937}     &\textcolor{blue}{.966}       & \textcolor{blue}{.957}     & \textcolor{blue}{.020}  
&   .944   & .964     & \textcolor{blue}{.967}      & \textcolor{blue}{ .022  }
&   .877    &.912     & .868      &  .054     \\

VSCode \cite{vscode} &CVPR 24
&\textcolor{blue}{.917}       &\textcolor{blue}{.954}       & .910      &-
&.869       & .894      & .830      &-
&.935       &.965      & .946      & -
&\textcolor{blue}{ .945 }     & \textcolor{red}{.971  }    & .957      & -
& \textcolor{blue}{.878 }     & .900      & .852      & -
\\
\hline
\textbf{JoNet (Ours)}                    &  -
&\textcolor{red}{.931} &\textcolor{red}{.958} &\textcolor{red}{.929}  &\textcolor{red}{.020}

&\textcolor{blue}{.872} &\textcolor{blue}{.895}      &\textcolor{blue}{.861}       & \textcolor{red}{.040}

&\textcolor{red}{.938} &\textcolor{red}{.967}  & .949 &\textcolor{red}{.019} 

&\textcolor{red}{.948} &\textcolor{blue}{.966}   & \textcolor{red}{.968}     &  \textcolor{red}{.020}

&\textcolor{red}{.887} &\textcolor{red}{.924}  &.885   & \textcolor{red}{.046}     \\ \hline

\end{tabular}}
\label{table:sodsota}
\end{table*}
\begin{table*}[t]
\centering
\caption{Quantitative comparison of our method with other SOTA methods on four COD benchmark datasets.}

\resizebox{\linewidth}{!}{\begin{tabular}{c|c|cccc|cccc|cccc|cccc}
\hline
\multirow{2}{*}{Method} & \multirow{2}{*}{Venue} 
& \multicolumn{4}{c|}{CAMO }                     
& \multicolumn{4}{c|}{CHAMELEON }                    
& \multicolumn{4}{c|}{COD10K}                  
& \multicolumn{4}{c}{NC4K} 
\\ \cline{3-18} 
&                      
&$S_{\alpha}\uparrow$ &$E_{\phi}\uparrow$  &$F_{\beta}^{\omega}\uparrow$  &\multicolumn{1}{c|}{$M \downarrow$}       
&$S_{\alpha}\uparrow$ &$E_{\phi}\uparrow$  &$F_{\beta}^{\omega}\uparrow$  &\multicolumn{1}{c|}{$M \downarrow$}     
&$S_{\alpha}\uparrow$ &$E_{\phi}\uparrow$  &$F_{\beta}^{\omega}\uparrow$  &\multicolumn{1}{c|}{$M \downarrow$}   
&$S_{\alpha}\uparrow$ &$E_{\phi}\uparrow$  &$F_{\beta}^{\omega}\uparrow$    & $M \downarrow$    
\\ \hline

SINet \cite{SINet}  & CVPR 20             & .751 & .771 & .606 & .100 & .869 & .891 & .740 & .044 & .771 & .806 & .551 &.051 & .810 & .873 & .772 & .057 \\

BGNet \cite{BGNet} &IJCAI 21      &  .812     &   .870    &   .749    & .073      &  .901     & .943      &  .850     & .027      & .831      &  .901     &  .722     & .033      &  .851     &    .907   &   .788    & .044      \\

SINet-V2 \cite{SINet-V2} &TPAMI 21   &  .820     & .882      &  .743     & .070     &  .888     &  .942     & .816      & .030       &  .815     & .887      &   .680    & .037     & .847      &  .903     & .769      &  .048     \\

UJSC \cite{UJSC} &CVPR 21     & .803      &  .853     &  .759     & .076      &  .894     &  .943     &  .848     & .030      &   .817    & .892      & .726      & .035     &  .842     &   .907    &  .771     &  .047     \\

LSR \cite{NC4K}  &CVPR 21    &  .793     &   .826    & .725      & .085      & .893      & .938      & .839      &.033    & .793      &  .868     &   .685    & .041    & .839      &  .883     & .779      &   .053    \\

BGSANet \cite{BGSANet} &AAAI 22 
&.796 &.851 &.768 &.079
&.895 &.946 &.851 &.027
&.818 &.891 &.723 &.034
&.841 &.897 &.805 &.048 \\

OSFormer \cite{OSformer} &ECCV 22
&.799 &.858 &.767 &.073
&.891 &.939 &.836 &.028
&.811 &.881 &.701 &.034
&.832 &.891 &.790 &.049 \\

ZoomNet \cite{ZoomNet}   &CVPR 22 &  .820     &  .892     &   .752   & .066     & .902      &  .958     & .845      & .023     &  .838     &   .911    &   .729    & .029      & .853      &  .912     &   .784    &  .043     \\

SegMaR \cite{SegMAR} &CVPR 22          &  .815     &  .872     &  .742     & .071    &  .906     & .954      & .860      &.025     &  .833     &  .895     & .724      & .033      &  .845     &  .892     & .793      &  .047     \\

FPNet \cite{FPNet1}  &MM 23              &  .852     &  .905    &  .806     & .056      & \textcolor{blue}{.914}     &  \textcolor{blue}{.961}     &  .856    & \textcolor{blue}{.022}      &  .850     & .913      &  .748   & .029    &  -     &  -     &   -    & -    \\

EVP \cite{EVP} &CVPR 23             & .846      &  .895     &  .777     & .059      &  .871     & .917      &    .795   & .036      &   .843    &  .907     &  .742     & .029    &  -     &   -    &  -     & -      \\

FEDER \cite{FEDER} &CVPR 23         &  .822     & .886      & .809    & .067      &   .907    &  \textcolor{red}{.964}     &  \textcolor{blue}{.874}     & .025     
&  .851     &  .917    & .752    & .028      &   .863    &.917     & .827  &  .042     \\

FSP \cite{FSP}  &CVPR 23            &.856     &  .899     &  .799     & .050     & .908      &  .942     &  .851     &.023     &.851     & .895      &  .735     & .026     &  .879    &  .915     &   .816    & \textcolor{blue}{.035}      \\ 
DINet \cite{DINet} & TMM 24
&.821 &.883 &.763 &.068
&- &- &- &-
&.832 &.914 &.780 &.031
&.856 &.919 &.839 &.043
\\
RISNet \cite{RISNet} &CVPR 24
&.870 &.922 &.827 &.050
&.909 &.955 &.871 &.023
&\textcolor{blue}{.870} &.921 &\textcolor{blue}{.781} &.027
&.882 &.925 &.834 &.036 \\

CamoFormer \cite{CamoFormer} &TPAMI 24
&\textcolor{blue}{.876} &\textcolor{red}{.935} &\textcolor{blue}{.832} &\textcolor{blue}{.043}
&- &- &- &-
&.862 &\textcolor{red}{.932} &.772 &\textcolor{blue}{.024}
&\textcolor{blue}{.888} &\textcolor{red}{.941} &\textcolor{blue}{.840} &\textcolor{red}{.031} \\

\hline

\textbf{JoNet (Ours)}      &-                    
& \textcolor{red}{.889}  &  \textcolor{blue}{.931}   &\textcolor{red}{.847}  &\textcolor{red}{.038}

& \textcolor{red}{.916}  &  \textcolor{blue}{.961}    &  \textcolor{red}{.876}   & \textcolor{red}{.020}

& \textcolor{red}{.871}  &  \textcolor{blue}{.925}    &\textcolor{red}{.783} &\textcolor{red}{.023} 

&\textcolor{red}{.895}  & \textcolor{blue}{.935}  & \textcolor{red}{.848}  & \textcolor{red}{.031} 
\\ \hline

\end{tabular}}
\label{table:codsota}

\end{table*}

We continue by reporting the performance of using SBSS and SCJoint alone. It can be clearly seen that when only SBSS is used for joint learning, the training data volume of the two tasks reaches a balance, and the network will not be dominated by one of the tasks. The performance of the COD task improves significantly compared to JT, but still falls short of IT, as the network is confused by the joint supervision of the two tasks at this time; the performance of SOD remains almost unchanged compared to JT, although the volume of SOD training data is reduced, its quality is improved. When only SCJoint is applied for joint learning, compared to IT, the performance of SOD slightly improves, but the performance of COD shows almost no improvement. The performance of both tasks does not decline compared to IT, indicating that their joint training does not confuse the network. This has addressed the first challenge we proposed, but due to the presence of the second challenge, not both tasks have benefited from joint learning.

Finally, we report the performance of simultaneously applying SBSS and SCJoint to joint learning, \textit{i.e.}, using our joint learning network JoNet. It can be observed that compared to IT, both tasks exhibit significant improvements in performance, truly benefiting from joint learning, which aligns with our expected goal. Compared to JT, although the improvement in the SOD task is not substantial, there is a significant enhancement in the COD task. Moreover, we achieve this without a significant increase in parameter count (DLM introduces only a negligible amount of parameters), and SBSS also reduces training time by nearly half (this is compared to training the two tasks separately, as we only used less than half of the SOD training data), achieving a win-win situation in terms of both performance and efficiency. These experiments and comparisons have proven the effectiveness of our proposed SBSS and SCJoint, especially their powerful capabilities when used together, enabling SOD and COD tasks to benefit from our joint learning network JoNet.

\vspace{1mm}
 \textbf{DLM location}. We discuss the performance impact of inserting DLM at different locations in \cref{tab:DLM_location}. It can be clearly seen that the best performance is obtained when DLM is only inserted into the decoder, which is also our default setting in this work. As long as DLM is inserted into the encoder part (corresponding to “encoder” and “encoder and decoder” in \cref{tab:DLM_location}), that is, learning the different distribution characteristics of the two tasks during the encoding process, it will bring devastating damage to the performance of the SOD task. It seems that at this time the network only has the ability to handle COD tasks. As for why it becomes like this, it is also a question worthy of our further investigation. But the experimental results have proved the correctness of our point of view, that is, the encoding process for the SOD and COD tasks is unified, while these two decoding processes have different distribution characteristics and require separate learning by the network.

We continue to discuss the impact of inserting DLM at different layers of the decoder. It can be clearly seen that when DLM is inserted into more and more layers of the decoder, the overall trend of network performance changes is gradually getting better. We believe this is because more DLMs make the decoder more capable of learning the characteristics of both tasks. Therefore, in order to get the best performance, we insert DLM into each Transformer block of the decoder. Due to the efficiency of DLMs, this approach does not significantly change the total number of network parameters and computational load. 

\vspace{1mm}
 \textbf{Sampling selection}. We report the results of sampling data with different rankings in \cref{tab:sbss}. According to the saliency ranking of image-GT pairs in $\mathcal{D}_{S}$ calculated by SBSS, we take the first K, last K, and randomly selected K image-GT pairs to form $\mathcal{D}_{S}^{sub}$ for joint training. It can be clearly seen that Top-K selection has the best performance, Bottom-K selection has the worst performance, and Random-K selection has a performance in between, which is consistent with our expectations, that is, the Top-K partial data quality is the best and Bottom-K is the worst, and the quality of training data has a greater impact on the network. In addition, we find that different SOD sample selections will affect the performance of the joint learning network JoNet on the COD task. We speculate that some low-quality samples may have brought wrong supervision signals to the network, affecting the segmentation ability of the network, which is a critical capability important to both tasks and then affects the network's learning of the commonalities.

It should be noted that our starting point is to filter out the low-quality image-GT pairs in the SOD dataset as shown in \cref{fig:2}, which account for a certain proportion of the SOD dataset. Although our method may change the distribution of the original training set, the experimental results prove that it is effective.

\vspace{1mm}
 \textbf{Impact of different data scale}. In the \cref{sec:intro}, we point out a major challenge of jointly training SOD and COD tasks, which is the imbalance in the sizes of the training sets for the two tasks. When the dataset size of one task is much larger than that of the other, simple joint training will be dominated by the task with the larger dataset size. Here, we report the performance of jointly training the network using different data ratios in \cref{tab:balance} to observe the impact of different proportions of training data on the performance of the two tasks. Note that our joint training here refers to simply mixing the data of the two tasks together for training, without using the SCJoint and SBSS strategies we proposed.

We fixedly use 4040 COD training images, and add $1000$, $2000$, $4040$, $8000$ and $10553$ randomly selected SOD training images respectively for joint training ($4040$ and $10553$ are the number of training images contained in the training sets of the two tasks respectively). It can be clearly observed that as the number of training images for SOD increases, \textit{i.e.}, the proportion of SOD data in the training data increases (approximately $20\%$, $30\%$, $50\%$, $65\%$ and 75\% respectively), the performance of the network on the SOD task improves while deteriorating on the COD task. This is consistent with our assumption that the training process of the network is dominated by the task with larger data sizes. Therefore, in this work, we propose the SBSS to address this challenge.

\vspace{1mm}
 \textbf{Comparison of SCJoint with other joint learning methods.} We compare our SCJoint with three other joint learning strategies mentioned in \cite{SENet, UJSC, vscode} in \cref{tab:joint}. We reproduce these three strategies on the baseline model SENet, and all use SBSS for fair comparison. It can be seen that the “decoder sharing” strategy is the most unsuccessful, and the performance of both tasks has dropped significantly compared to IT. The remaining three methods can effectively complete joint learning,  at least without causing performance degradation. Among them, our proposed method has the most obvious improvement. In addition, the “encoder sharing” and “decoder sharing” methods both bring more additional parameters to the network, about $30\%$ and $70\%$  more than baseline. Our proposed SCJoint and “prompt learning” method only introduce negligible additional parameters. These prove the effectiveness and efficiency of the proposed SCJoint. 
In addition, we experimented with using LayerNorm to replace DLM, and it can be observed (see the last two rows in \cref{tab:joint}) that DLM achieves better experimental results in practice.

\begin{table}[t]
\centering
\caption{Introduction to the dataset used in the Shadow detection task.}
\begin{tabular}{c|c|c|c}
\hline
Task                                                                        & Dataset & Train & Test \\ \hline
\multirow{2}{*}{\begin{tabular}[c]{@{}c@{}}shadow\\ detection\end{tabular}} & ISTD \cite{ISTD}    & 1330  & 540  \\ \cline{2-4} 
    & SBU \cite{SBU}    & 4089  & 638  \\ \hline
\end{tabular}
\label{tab:sd}
\end{table}

\begin{table*}[t]
\centering
\caption{Apply SCJoint to joint learning for more tasks. We select DUTS (SOD dataset) \cite{DUTS} , CAMO (COD dataset) \cite{CAMO} and ISTD (shadow detection dataset) \cite{ISTD} to evaluate the performance of the joint learning network. “IT” indicates independent training, that is, training on the respective task training sets.  “JT” indicates normal joint learning, which simply mixes the three data together for training.  During joint training, we randomly select $4040$ training images from the training set of each task to ensure data balance.}
\resizebox{\linewidth}{!}{\begin{tabular}{cccccccccccccc}
\hline
\multicolumn{1}{c|}{\multirow{2}{*}{Settings}} 
& \multicolumn{4}{c|}{DUTS (SOD)}                    
& \multicolumn{4}{c|}{CAMO (COD)}                    
& \multicolumn{4}{c}{ISTD (SD)}                  \\ \cline{2-13}
\multicolumn{1}{c|}{}              

&$S_{\alpha}\uparrow$ &$E_{\phi}\uparrow$  &$F_{\beta}^{m}\uparrow$  &\multicolumn{1}{c|}{$M \downarrow$}       

&$S_{\alpha}\uparrow$ &$E_{\phi}\uparrow$  &$F_{\beta}^{m}\uparrow$  &\multicolumn{1}{c|}{$M \downarrow$}     


&$S_{\alpha}\uparrow$ &$E_{\phi}\uparrow$  &$F_{\beta}^{\omega}\uparrow$    & $M \downarrow$     \\ \hline

\multicolumn{1}{l|}{IT}               
&.921  &.953  &.921  & \multicolumn{1}{c|}{.024}  
&.875  &.916  &.832  & \multicolumn{1}{c|}{.043}  
&.944  &.966  &.926  &.014  \\ \hline

\multicolumn{1}{l|}{JT (SOD+COD+SD)}              
&.868  &.907  &.828  & \multicolumn{1}{c|}{.041}  
&.854 &.900  &.799  & \multicolumn{1}{c|}{.051}  
&.942  &.962  &.921  & .018 \\  \hline

\multicolumn{1}{l|}{JT (SOD+COD+SD) with SCJoint}              
&\textbf{.926}  &\textbf{.953}  &\textbf{.923}  & \multicolumn{1}{c|}{\textbf{.021}}  
&\textbf{.883}  &\textbf{.922}  &\textbf{.843}  & \multicolumn{1}{c|}{\textbf{.042}}  
&\textbf{.948}  &\textbf{.972}  &\textbf{.934}  &\textbf{.012}  \\ 





\hline

\end{tabular}}
\label{tab:more tasks}
\end{table*}

\begin{figure}[t]
      \centering
      \includegraphics[width=0.9\linewidth]{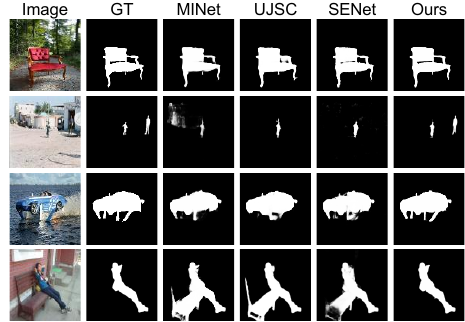}
      \caption{Visual comparison of our saliency predictions with the SOTA methods. }
      \label{fig:sod_visualization}
\end{figure}

\begin{figure}[t]
      \centering
      \includegraphics[width=0.9\linewidth]{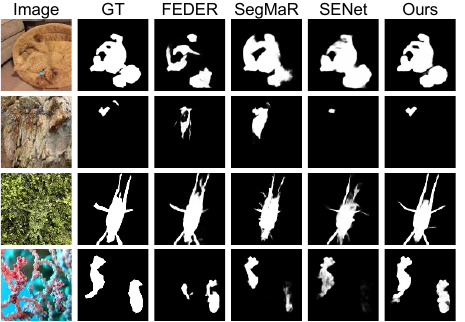}
      \caption{Visual comparison of our camouflage predictions with the SOTA methods. }
      \label{fig:cod_visualization}
\end{figure}

\subsection{Comparison with SOTA Methods}
 \textbf{Performance comparison on SOD}. \cref{table:sodsota} summarizes the quantitative results of our JoNet against 17 competitors on five
 challenging SOD benchmark datasets under four evaluation metrics. It can be clearly seen that among the 20 comparisons, our JoNet achieves 14 best results and 4 sub-optimal results, which proves the strong performance of our method. In addition, we have a large lead on the DUTS and PASCAL-S datasets, improving the indicators a lot compared to the second place. Competitors include joint learning generalist models such as EVP \cite{EVP} and UJSC \cite{UJSC}, as well as task-specific specialist models such as VST \cite{VST}, the performance of our method exceeds these.

\vspace{1mm}
\textbf{Performance comparison on COD}. \cref{table:codsota} summarizes the quantitative results of our JoNet against 16 competitors on four
 challenging COD benchmark datasets under four evaluation metrics. It can be clearly seen that among the 16 comparisons, our JoNet achieves 12 best results and 4 sub-optimal results, which proves the strong performance of our method. In addition, we have a large lead on the CAMO, COD10K and NC4K datasets, since the difficulty of the CHAMELEON dataset is low, we are not far ahead.

 Although our joint training model JoNet performs well on both SOD and COD tasks, it is undeniable that our excellent performance mainly comes from our powerful baseline model SENet \cite{SENet}. However, the focus of our paper is to overcome the challenges in joint learning of SOD and COD tasks to make the two tasks benefit from our joint training, the performance of our joint learning model JoNet is still improved compared to the baseline model trained separately, see \cref{tab:efficacy}.

\begin{figure}[t]
      \centering
      \includegraphics[width=0.9\linewidth]{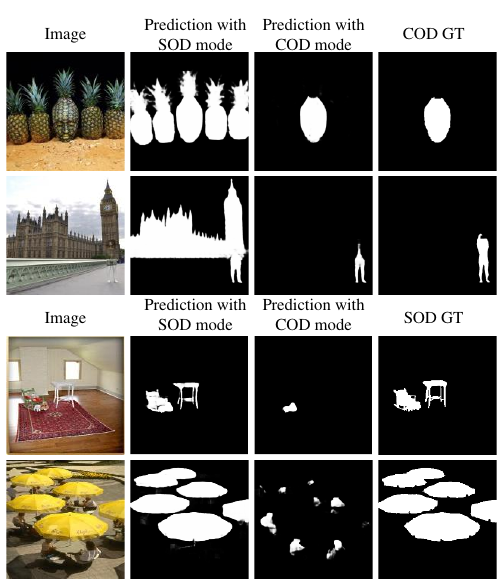}
      \caption{Visualization of prediction maps obtained using different modes of inference for the same image. We select two images from the SOD and COD testing sets respectively, so they only have one kind of GT, the images from SOD only have the corresponding SOD GT and the images from COD only have the corresponding COD GT. It can be clearly seen that the predictions of the mode corresponding to GT are very similar to GT, while the predictions of the other mode are very different. Please zoom in for more details.}
      \label{fig:qualitative}
\end{figure}

\begin{figure}[t]
      \centering
      \includegraphics[width=0.8\linewidth]{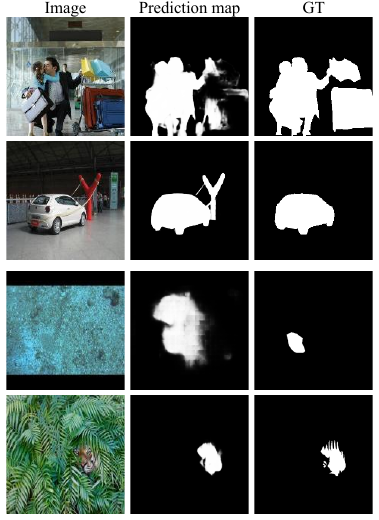}
      \caption{Visualization of failure cases. The top two rows are samples of the
SOD task, and the bottom two rows are samples of the COD task.}
      \label{fig:failure}
\end{figure}

\begin{figure}[t]
      \centering
      \includegraphics[width=0.9\linewidth]{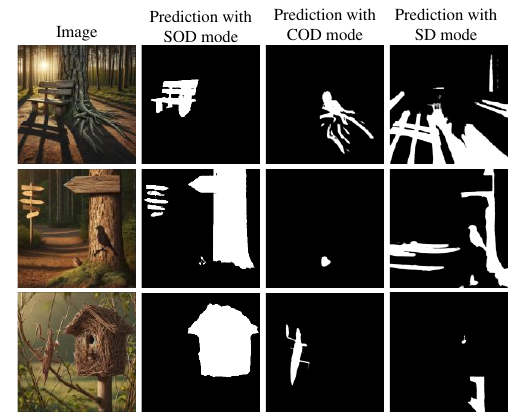}
      \caption{Visualization of prediction maps obtained using different modes of inference for the same image. Given an image, our network is able to simultaneously discover “salient”, “camouflaged” and “shadow”. Note that since the testing set we used rarely contains images with salient objects, camouflaged objects and shadows at the same time, the images we show are generated using GPT4 \cite{gpt4}, so they have no GT. Please zoom in for more details.}
      \label{fig:three_mode}
\end{figure}

\subsection{Qualitative Visualization and Analysis}

We present visual comparisons of JoNet with several SOTA methods for the SOD and COD tasks in \cref{fig:sod_visualization} and \cref{fig:cod_visualization}, respectively. As shown in the figures, JoNet achieves more accurate target localization and produces segmentation maps with more complete structures and sharper, more accurate edges that are closer to the ground truth (GT). These results demonstrate that JoNet performs well on both tasks simultaneously, meeting our expected objectives.

We also illustrate the ability of our joint learning network JoNet to simultaneously capture both “salient” and “camouflaged” in \cref{fig:qualitative}. It can be seen that for the same input image, we have two modes for inference (using different DLMs). When using the SOD mode, the network focuses on the most salient object in the image and provides its binary segmentation mask. When using the COD mode, the network identifies the hidden camouflaged objects in the image and provides the binary mask. It can be seen that the prediction maps given by the network are of high quality and are basically consistent with our human judgment, indicating that the network has mastered strong segmentation capabilities. This requires the network to have strong perception capabilities for pictures, including the perception of high-level semantics of pictures and the grasp of underlying details, including colors, textures, edges, \textit{etc.},all of which are very important for both tasks. In addition, the prediction results of the two modes are very different, which also shows that the network has indeed learned the different characteristics of the two tasks.  We do not set up any supervision to deliberately amplify these differences. This is learned spontaneously by the network, which also proves the success of our joint learning method. This grasp of the commonalities and properties of the two tasks is consistent with our intended goals.

\subsection{Failure Case and Analysis}
We present some failure cases in \cref{fig:failure} to analyze
the limitations of the proposed method. Although our JoNet can perform both SOD and COD tasks simultaneously and has achieved promising quantitative results on multiple datasets, it can be observed from the figure that the segmentation performance is still not satisfactory when dealing with some challenging scenarios. Errors such as inaccurate target localization, incomplete segmentation, and blurred edges may occur. This indicates that JoNet’s understanding of the overall target regions in complex scenes remains insufficient, and its perception capability for complex scenes needs further improvement. This is also one of the key directions we plan to focus on in future work.

\subsection{The Potential of SCJoint}
We apply the proposed joint learning strategy SCJoint to additional binary segmentation tasks to explore its potential. We find another binary segmentation task, shadow detection (SD), which aims to find the shadow parts in the image. We try to put it together with SOD and COD tasks for joint training to explore whether the proposed SCJoint has the ability to be applied to joint training of more tasks. We introduce the datasets commonly used by SD task in \cref{tab:sd}.

As shown in \cref{tab:more tasks}, it can be clearly seen that normal joint training will cause the performance of the three tasks to decline compared to IT, but joint training using the SCjoint strategy (set up separate DLMs for each of the three tasks in the network) will make the three tasks benefit from joint training, namely, the performance will be improved compared to IT. At this time, the network has the ability to discover “salient”, “camouflaged” and “shadow” at the same time, as shown in \cref{fig:three_mode}. This proves that \textbf{our proposed SCJoint has the potential to be applied to more multi-task joint learning}, and is not limited to SOD and COD tasks. It is also worth noting that the images in \cref{fig:three_mode} are generated by GPT4, which means that they belong to a different distribution from our training sets and are out-of-distribution (OOD) samples, which to a certain extent proves the generalization ability of our method.

\section{Conclusion}
\label{sec:conclusion}
In this work, we propose SCJoint and SBSS to address two challenges encountered in joint learning of SOD and COD tasks. SBSS helps us balance the dataset sizes of the two tasks, improve the training set quality of the SOD task, and greatly shorten the training time, thus improving efficiency. SCJoint beautifully learns the characteristics of the two tasks at the cost of introducing a very small amount of learnable parameters, so that our joint learning network has the ability to capture “salient” and “camouflaged” at the same time. SCJoint and SBSS enable SOD and COD tasks to benefit from joint learning and achieve our expected goals.

\bibliographystyle{IEEEtran}
\bibliography{main}
\end{document}